\documentclass[a4paper,10pt]{article}

\usepackage{amsfonts}
\usepackage{amssymb}
\usepackage{latexsym}
\usepackage{graphicx}

\makeatletter
\renewcommand\@openbib@code{%
     \advance\leftmargin  \z@ 
      \itemindent \z@
     \parsep -0.8ex
     }
\makeatother

\makeatletter
\renewcommand\section{\@startsection {section}{1}{\z@}%
                                   {-3.5ex \@plus -1ex \@minus -.2ex}%
                                   {2.3ex \@plus.2ex}%
                                   {\large\bfseries}}
\makeatother

\makeatletter
\renewcommand\subsection{\@startsection {subsection}{1}{\z@}%
                                   {-3.5ex \@plus -1ex \@minus -.2ex}%
                                   {2.3ex \@plus.2ex}%
                                   {\normalsize\bfseries}}
\makeatother

\begin{document}
\pagestyle{empty}

\begin{center}
{\bf \Large MOBILE AUGMENTED REALITY APPLICATIONS}
\end{center}

\smallskip
\begin{center}
{\large 
David Prochazka, Michael Stencl, Ondrej Popelka, Jiri Stastny
}
\end{center}

\smallskip
\begin{center}
Mendel University in Brno\\
Faculty of Business and Economics\\
Department of Informatics\\
Zemedelska 1, 61300 Brno\\
Czech Republic\\
david.prochazka@mendelu.cz
\end{center}

\bigskip
\noindent Abstract: \textit{Augmented reality have undergone considerable improvement in past years. Many special techniques and hardware devices were developed, but the crucial breakthrough came with  the spread of intelligent mobile phones. This enabled mass spread of augmented reality applications. However mobile devices have limited hardware capabilities, which narrows down the methods usable for scene analysis. In this article we propose an augmented reality application which is using cloud computing to enable using of more complex computational methods such as neural networks. Our goal is to create an affordable augmented reality application suitable which will help car designers in by 'virtualizing' car modifications.}

\vspace*{10pt} \noindent Keywords: \textit{Augmented Reality, Image Processing, Marker, Neural Network, Mobile Devices}

\bigskip
\section{Introduction}
Augmented reality (AR) is scientific field well known for many decades. As early as in 1993 one issue of the \textit{Communications of the ACM} was dedicated to the new emerging field of augmented reality and related ubiquitous computing. One of the significant contributors was Mark Weiser from Xerox Palo Alto laboratories. In \cite{1} and few related articles the key problems for this field were formulated. A significant number of them is still unsolved and augmented reality is still far from being a common tool.

Almost two decades later, in 2002, Billinghurst and Kato published article \cite{2} that summarized state-of-the-art of the collaborative augmented reality and in a broader view the whole augmented reality. The gist of this article is formulated in the conclusion: "Despite early promising results, a lot of research work needs to be done before collaborative AR interfaces are as well-understood as traditional telecommunication technology." This description could have been applied on most of the AR applications. One of the obvious reasons was the necessity of obtaining special hardware -- helmets, binoculars, etc.

\begin{figure}[htb]
\begin{center}
\includegraphics[scale=0.5]{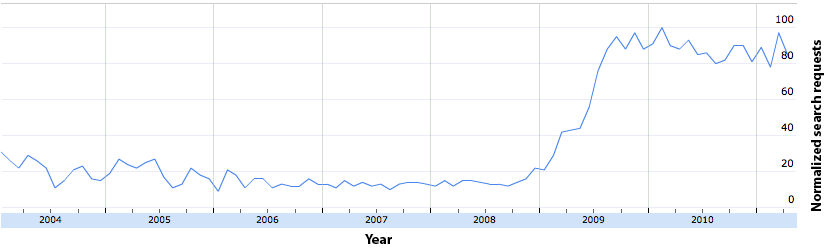}
\end{center}
\caption{Google search request for augmented reality between years 2004 and 2011. The \textit{y} axis shows search requests normalised to range from 0 to 100. This number represents a fraction of highest search activity. The normalisation method can be found on \textit{Google Search Insights} web page.}
\label{google}
\end{figure}

However, with the introduction of cell phones equipped with camera a completely new tool evolved. Now, there exists a cheap common device which is able to present a scene composed of real-word image and different artificial objects. Such presentation is also very natural. After few years, it is obvious that adoption of \emph{mobile augmented reality} by common users is much faster than the adoption of previous applications (see \cite{3}, \cite{4} and many others). This trend is clearly presented also on Fig. \ref{google} that presents normalised global search requests for "augmented reality" keywords. The exact reason for the boom about the year 2009 will be explained later. 

In spite of the significant amount of research in the field of AR, there is still a discussion about optimal approach for recognition of the real-world scene -- in other words: where the user is and what is before him. The following section briefly describes frequently used approaches of solving this problem. We will focus especially on mobile AR applications.

\subsection{Motivation}

In this article we present the concept of augmented reality application designed to aid car designers. Car designers struggle with many problems when creating new car designs or when customizing current ones.

One of the objectives is to reuse as many standard parts as possible while still producing interesting new designs. In order to test and actually see different parts  (head-light, wheel, spoiler, etc.) a designer has to modify the physical car body, which is costly and slow. Our current work is focused on development of applications which would enable the designer to virtually replace car parts on a real car or physical model. We already presented a solution which allows to augment the of the prototype camera image on large screen projection by virtual objects (see \cite{18}). This solution is based on artificial markers. 


Our current goal is to develop an affordable mobile application which will require no special equipment (such as markers) and still provide reasonable user experience when presenting the vehicle modifications. In this article we describe a concept of an application which will provide a quick preview of an existing car and a virtual modification to it. Purpose of this article is to present our current approach and receive a feedback on proposed solution based on soft computing.

\section{Approaches for Scene Identification}

Augmented reality applications can be divided in two main groups according to the device type used -- applications for \emph{see-through devices} and applications for \emph{devices with video composition} of the scene.

See-through devices are used mostly for special purposes. Well-known example is the see-through display in a fighter cockpit visualising position and speed of enemy aircraft. Another widely spread application is the projection of navigation information on the front windshield in a car. These applications are accompanied by a number of special purpose solutions based on see-through head mounted displays (see \cite{16}). The basic principle of these devices is that user is able to see real world directly through a see-through layer on which the digital information is projected.

The group of devices based on the video composition also includes head mounted displays (HMD). In this case the user is able to see just a digital record of the real world scene extended with the digital information. Together with the HMD, virtually all mainstream AR applications on tablets and cell phones are based on this principle.

In depth comparison of mentioned approaches including discussion about the immersibility of different solutions could be found e.~g. in \cite{4}. 



\subsection{Mainstream Application: Personal Navigation}

\begin{figure}[htb]
\begin{center}
\includegraphics[scale=0.4]{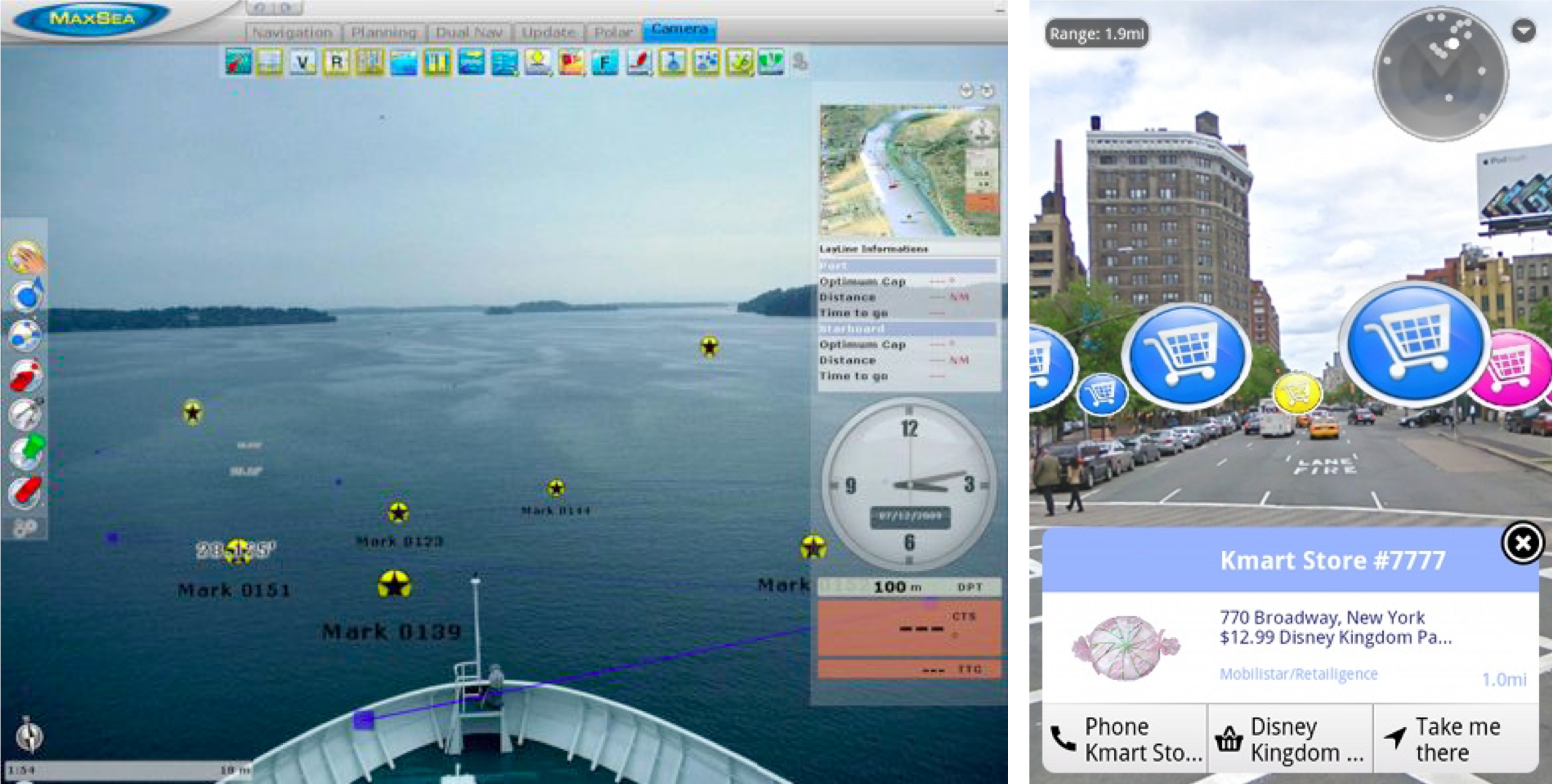}
\end{center}
\caption{Experimental maritime navigation and \textit{Layar} application on a cell phone}
\label{maritime}
\end{figure}

These applications could be frequently described as \emph{location based services}. They are usually not based on image analysis, but on input from different sensors (compass, GPS, RFID, etc.). Image from the camera is just used as a background of the final scene. An example of such approach is the experimental augmented reality platform for assisted maritime navigation described in \cite{5} (see left part of Fig. \ref{maritime}). A number of mainstream augmented reality applications such as successful \textit{Layar}\footnote{http://www.layar.com/} is also based on this principle. Both mentioned projects are using GPS and compass for location of the device. Using this information, they are embedding specific contextual information in the real-world image (route to the destination, nearby shops, points of interest, etc.).

These relatively simple applications are one of the reasons for augmented reality interest advancement around year 2009. Many contemporary cell phones were equipped with GPS, compass, camera and suitable display, therefore first mainstream applications such as the mentioned \textit{Layar} could emerge.

The limitation of this approach is obvious. The application has no information about any objects shown inside the camera picture. Therefore the presented information could be connected just to the direction and position of the device, not to a specific object. This group of applications usually includes different types of personal navigations.

On the other hand, a significant advantage of this approach is its' simplicity. It is not necessary to make any complex calculations. This is very important feature, since there are no special requirements on the hardware and also battery performance is not affected.

\subsection{Approaches Based on Image Processing}

Another group of augmented reality applications is based on image analysis. Image given by the camera is processed by the application engine, objects are identified and contextual information is embedded into the image and linked to the appropriate object. Especially devices based on video composition allow almost seamless overlaying of real and virtual objects. The following part of the article will focus especially on common mobile devices which are using this technique.

The adoption of mobile devices triggered a breakthrough of AR. However, some significant limitations must be overcome -- limited computation performance, memory, network connectivity, etc. Image processing techniques used must comply with these limitations.

Image processing solutions for AR can further be divided into two groups. The first group is focused on \emph{identification of artificial objects}. Usage of artificial objects (markers) significantly simplifies the image processing task. This object is usually easily found in the scene and could be clearly mathematically described. Due to this simplification, robust and fast algorithms could be developed. Example of an application based on this principle is mobile AR game \textit{ARhrrrr!} described in \cite{3} or scientific project such as \cite{6}. 

The other group is dealing with \emph{identification of natural objects} within the image (specific buildings, artefacts, etc.). In these cases are used frequently methods such as SURF. However, implementation of these methods on mobile devices is difficult (see \cite{17}). Because of this, many solution are simplifying this task back to the detection of simple elements. An example of such application is the cultural heritage browser described in \cite{9}. This solution is based on hybrid principle: general location is given by GPS and compass and exact objects in the scene are found by image analysis. These hybrid methods are also promising solution for many other applications -- e.g. logistics. Mobile spatial AR could significantly improve many logistical processes (see \cite{19}). 

 Soft computing methods are not used because of their complexness.

\section{Object Identification Technique for Artificial Objects}

\begin{figure}[hbt]
\begin{center}
\includegraphics[scale=0.45]{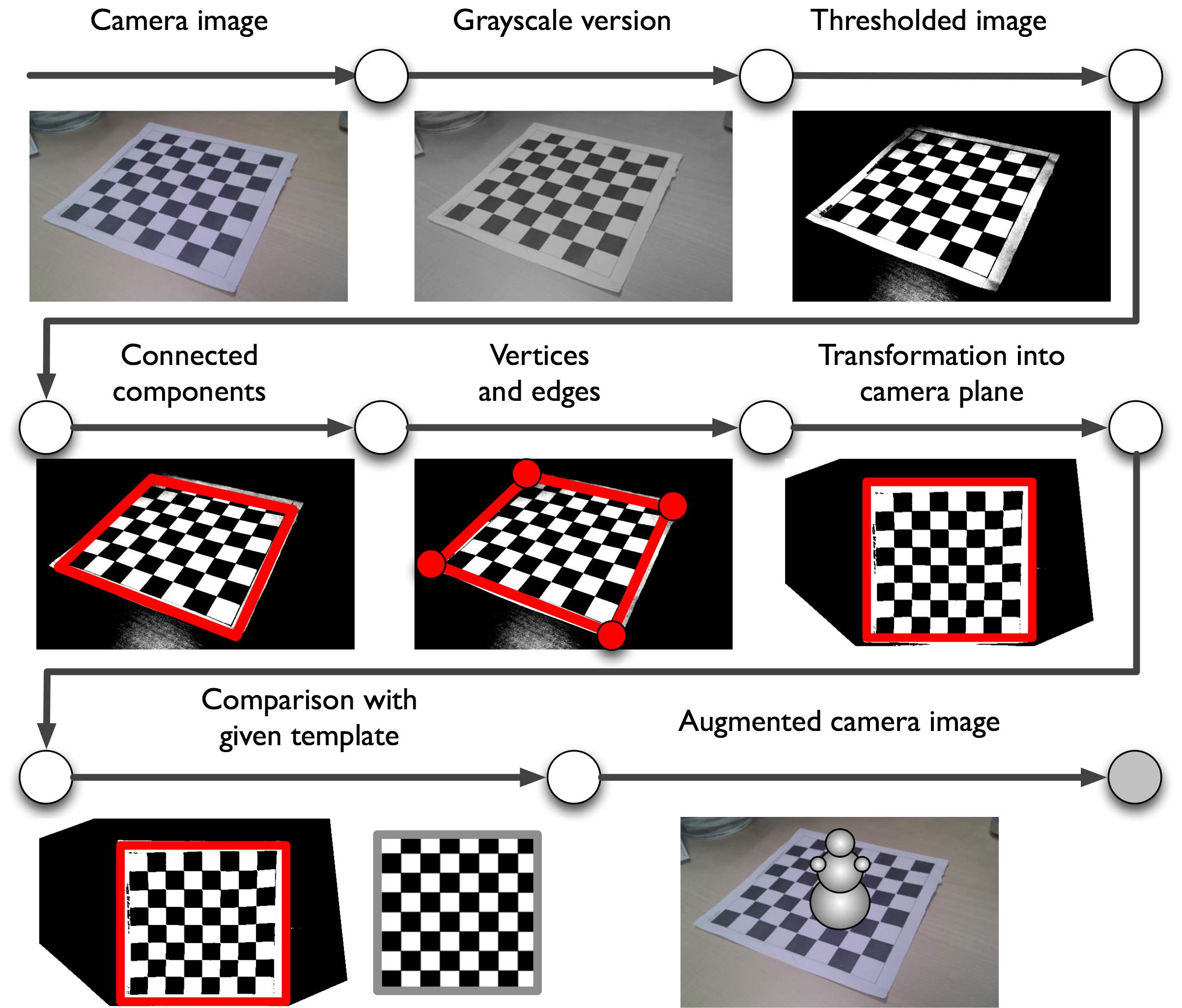}
\end{center}
\caption{General structure of a common augmented reality application based on artificial markers.}
\label{arscheme}
\end{figure}

The process of the object identification can be generally described by a set of operations outlined on Fig. \ref{arscheme}. Specific implementation of these steps can partially differ.

First three steps can be described as an \emph{image preprocessing}. In this phase, the image is usually converted into a grayscale version and then thresholded into binary image. Obviously there is a number of approaches designed for specific environments. Most significant problem in this phase is dealing with different light conditions in various locations. Therefore it is not possible to use a single threshold value for all light conditions. For example the paper \cite{10} is dealing with this problem and outlines the principle of enhanced algorithm for dynamic thresholding.

The following steps are the image analysis itself. The goal of this step is to decompose the image into a set of objects described by their vertices and edges. The application of the \emph{morphological analysis} method is usual solution. The result is a tree or list of image elements. At this moment it is possible to find all objects which fulfil some specific criterion (e.g. all potential squares given by four edges).

Further image analysis partially differs according to the nature of the application and what kind of objects should be identified. Common approach in AR is usage of different square markers. Advantage of a square marker is that it is extremely simple to find all such potential markers in the tree or list given by the morphological analysis. Each potential marker is transformed into the camera plane (step 5 on Fig. \ref{arscheme}.) and compared with predefined templates. Description of this transformation can be found in previously mentioned \cite{6}.

Another solution is an hybrid approach. It is possible to partially identify the position and orientation of the device using GPS and compass and match the scene exactly using simplified image analysis. This approach is used in \cite{9}. Image analysis used comprises of identification of specific shapes that should be found in a given location -- e.g. rectangle of a show case in the museum. From the point of view of image analysis this approach is similar to the previous case.

It is obvious that these techniques are not suitable for identification of complex structures such as buildings. In this case it is again necessary to identify some clearly specified features within the complex objects in order to simplify the image analysis. 

\subsection{Template Matching}

All mentioned techniques up to this point had similar basis: a region of interest was found in the preprocessed image. Then it is necessary to identify the content of the region (step 7 on Fig. \ref{arscheme}.). This step is completely different depending on the method used.

Frequently used method is based on calculation of correlation between image matrix of the template (something that we expect) and image matrix of the selected region of camera (part of the processed image). Several different implementations are described in \cite{12} in detail. In the well-known image processing library \textit{OpenCV} this step is represented by the \textit{Match Template} function.

The advantage of this approach is the ability to work with just a single original template. The main disadvantage is that it is necessary to make separate correlation computations for all templates up to the point when an appropriate template is found. This can introduce a significant computational load on the mobile device.

\begin{figure}[hbt]
\begin{center}
\includegraphics[scale=0.30]{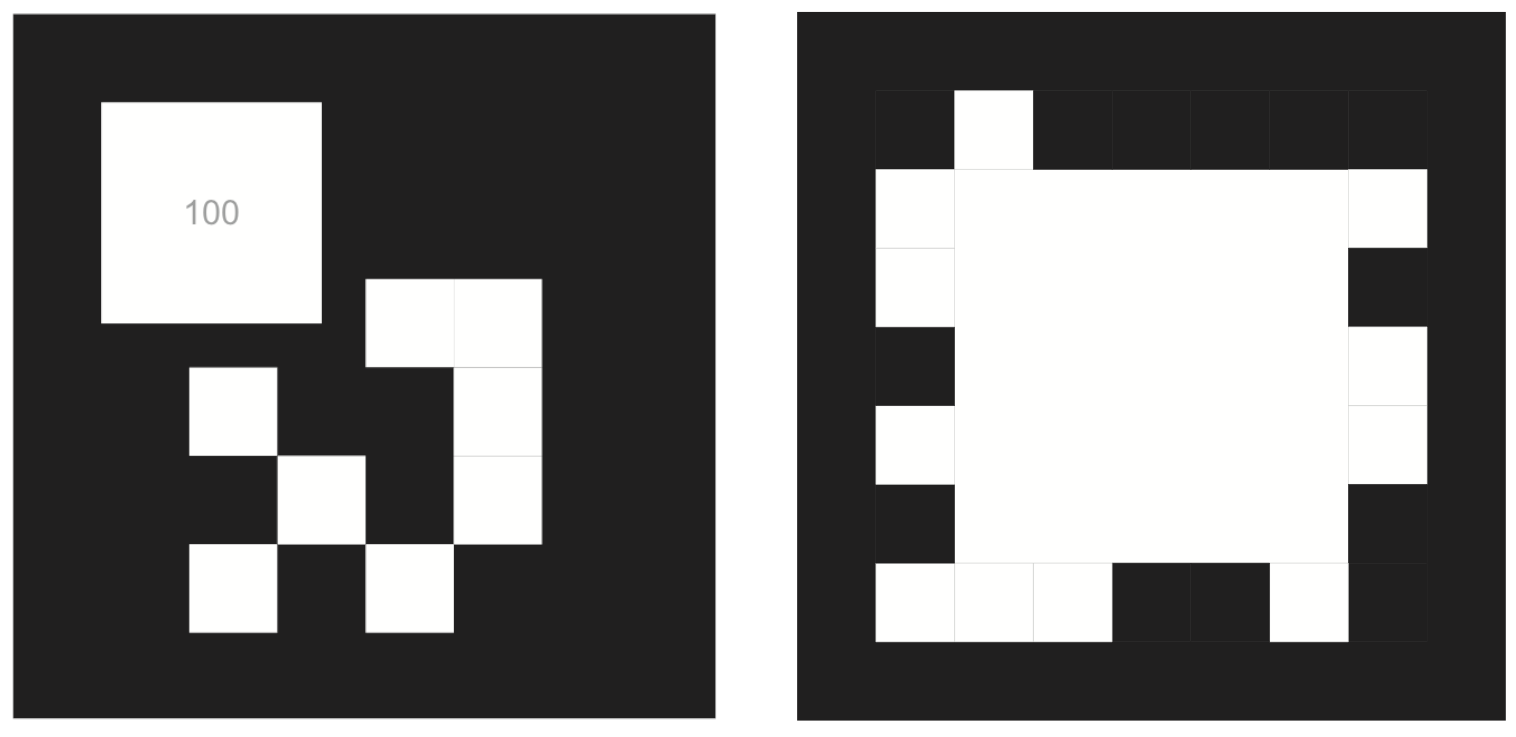}
\end{center}
\caption{Example of two artificial markers based on Golay error correction code.}
\label{golaymarker}
\end{figure}

From this point of view, it is much more suitable to use another technique to generate markers. For example a technique such as Golay error correction coding algorithm can be used. Markers based on Golay algorithm principle are just encoding a binary number into a set of black and white squares (see Fig. \ref{golaymarker}). Therefore it is not necessary to compare the potential marker with any template. The algorithm just identifies whether a given position is black or white color and transforms this information into 0 or 1 in a binary number. If the resulting number is not present in the list of defined markers, then the required object is found.

Image processing, especially the part of matching the camera image with a given template, requires a significant hardware resources. Currently used approaches are usually solving this problem by replacement of advanced image processing techniques by the input from GPS, compass, etc. Anothe option is using a special type of markes -- such as the described Golay algorithm is potentially much faster than template matching. Especially when the application is used with a large number of markers. However, it can not be used for identification of natural objects (unlimited markers would be needed), also it is not entirely suitable for our application as it would require the user to put markers on the automobile parts. This would disturb user experience.

\section{Mobile Application for Augmented Car Design}
\label{app-proposal}
In the optimal operations of the application, the user will use his mobile device (any sort of tablet) to 'look' at a real vehicle or it's physical model. User will then select part of the vehicle to be replaced. The application will connect to on-line part database (store) and allow the user to virtually exchange the part. The user will be allowed to adjust new part on the car. Important premises in designing the application are:
\begin{itemize}
\item The application should not require any setup except internet connection to server with part database.
\item It is not important to exactly identify the object beeing replaced, since it is being replaced.
\item It is only important to identify type of the object (wheel, head-light, spoiler, etc.).
\item The positioning of the overlay is not critical, since user is allowed to adjust the overlay.
\item Artificial markers cannot be used. Whole replacement must be done directly on mobile device.
\end{itemize}

As mentioned above, in the first step the user needs to select a car part, which will be replaced. This can be basically done in three ways. Either the user can select a single point of the part or area of the screen in which the part is located. These two approaches might be problematic since they can lead to ambiguity (some parts may be very close to each other or overlap). The third approach is to analyze the current picture first, and identify all possible car parts which can be replaced from the parts database. This seems a better approach since the user is apriori limited to replace only those parts which can be replaced (i.e. there are alternative parts in database) and also adds to user experience (eliminates possible inaccuracy in user selection).

The second step involves offering the user the list of possible replacements. For example, the user will request replacement of front wheel, so he will be offered a list of front wheels from the parts catalogue. By selecting a specific front wheel, it's image will be inserted into the car view. It is important to note, that the parts database contains 3D models of all parts, thus any view can be generated. The user can further adjust the overlay position and size manually.

\section{Detection of Complex Structures}

In the first step, it is necessary to identify a single object (car part) in the scene, or preferably identify all objects (car parts), which can be identified. In this case it is necessary to use some method for image recognition, specifically recognition of real objects. However, the task is not typical. Usualy, it is required that a specific object is identified -- for example face recognition, license plate recognition, etc. What we need in our application is identification of similar objects or identification of object types. Although there exists a database of known car parts, it can easily happen that the real part on the real vehicle is not in the database. Still it is neccessary to recognize what type of part it is (e.g. head-light) and again, it is not important what precise part it is (i.e. it does not matter wheter it is head-light model \textit{X} or model \textit{Y}). Therefore, it is required, that the image recognition method is capable of generalization, and actual precision is not important. For this reason we use neural network over known numeric methods such as previously mentioned SURF.

\subsection{Image recognition using neural networks}

Recognition with the aid of neural network algorithm is suitable where high-speed classification with randomly rotated objects is required and where we need to tolerate some differences between learned etalons and classified objects. This method can recognize objects with considerably modified shapes but it may identify incorrectly objects of similar shape.

In article \cite{20} several methods for object recongnition are compared, out of which the Multi Layer Perceptron neural network (MLP) algorithm prooved to be the fastest. The Multi Layer Perceptron neural network can be trained with different learning algorithms to provide better results. In this case the back-propagation learning algorithm achieved the best results \cite{24}. In this experiment different manufactured parts including similar objects had to be recognized in a technological scene.

\begin{figure}[hbt]
\begin{center}
\includegraphics[scale=0.3]{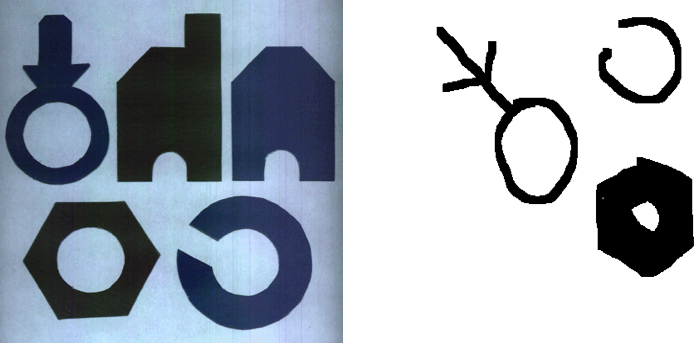}
\end{center}
\caption{Set of objects used as patterns and set of testing objects to recognize.}
\label{patternset}
\end{figure}

The real technological scene for object classification was simulated by digitizing five selected objects shown on the left side of Fig.~\ref{patternset}. For this purpose, two-dimensional images of three-dimensional objects were prepared. The aim was to test such objects that resemble two-dimensional images of real objects. The choice of objects of similar shape was also intentional. Each pattern was described by a flag vector. For the description of objects (i.\,e. on Fig.~\ref{patternset}) using this method, 70 symptomatic vectors were used that originated from the centre of gravity and directed to object edges. Flag vectors have been submitted to network and arm lengths have been transformed to values from interval $< 0, 1 >$.

After successful learning, the test set of several objects (i.\,e. right side of the Fig. \ref{patternset}) was given to the learned Multi Layer Perceptron neural network. The rate of success was examined by performing recognition on all technological scenes with edge detectors given in the comparison of edge detector speeds. For each edge detector new reference standards were prepared from the edges detected in the initial scene. The results of the simulation experiments show that the Multi Layer Perceptron neural network with the Back-Propagation algorithm (Fig.~\ref{patternset}) and the Levenshtein algorithm are very promising for object recognition of technological scenes for the use of industrial robots control \cite{21}. The Multi Layer Perceptron neural network shows good generalization abilities, however the flag vectors used require that the geometry of the object (or at least it's centre of gravity) is known.

An improved method is used in the article \cite{28} which describes using MLP neural network for road traffic sign recognition (TSR). Solving this problem also requires an algorithm which achieves greater generalization, while the precision is not that important. The authors used a MLP neural network for sign classificiation, however the input for the neural network was different. First, the SIFT (Scale-invariant feature transform) and SURF (Speeded-Up Robust Features) feature vectors were computed. The MLP neural network had 128 input neurons which corrseponded to the dimensionality of SIFT descriptor. The SIFT vectors were then fed to the neural network which performed the actual classification. The authors compared both SIFT and SURF feature vectors as input, concluding that SIFT feature vectors performed slightly better. This is an important result supporting the feasibility of the solution. Unlike flag vectors which rely on known geometry of the image, the SIFT descriptor can be computed on any type of image regardless whether the geometry is known or not.

\subsection{Cloud Computing}

The usage of neural network implies two problems. First it needs to be trained -- for this we use the on-line database of car part models. Second it is an algorithm with considerable computational complexity. For both these reasons it is not feasible to implement the neural network inside the mobile device. Solution of this problem can be usage of cloud computing principle. 

Proof of this concept is a well-know image processing application \textit{Google Goggles}\footnote{http://www.google.com/mobile/goggles/}. This application easily allows to identify many graphical objects from paintings to book titles. Client side of the application makes just basic analysis of the input and in most cases sends the image to a service for comparison with known images.

There is a number of significant advantages of this approach. The first and probably most important is that much more templates (neural networks) can be stored in the cloud service than in any device (mobile or desktop). Even more it is not necessary to update the template database on the client side. The other important advantage is, that much more advanced image processing algorithms can be used in a service than on common mobile devices. It is not necessary to adjust methods for mobile devices. Moreover, the server side is easily scalable, therefore performance of the solution can be easily increased.

Obvious disadvantage of this approach is the delay given by the communication. However a computation speed-up could compensate it. Also it is not necessary to query the server with every single movement of the mobile device. Local device sensors such as the accelerometer or compass, can be used to adjust the position of the overlay and respond to movement of the image. Also it is not necessary to re-match the objects in the scene, if the scene does not change drastically. Similar application of this concept -- optical character recognition -- is outlined in \cite{14}. As shown on Fig. 2 in the article, time necessary for remote procedure call waiting for the response is in almost all cases lesser than the computation on the device. Further discussed are also economical aspects of this solution. Successful experiments with this architecture are also described in \cite{15}.

\begin{figure}[hbt]
\begin{center}
\includegraphics[scale=0.55]{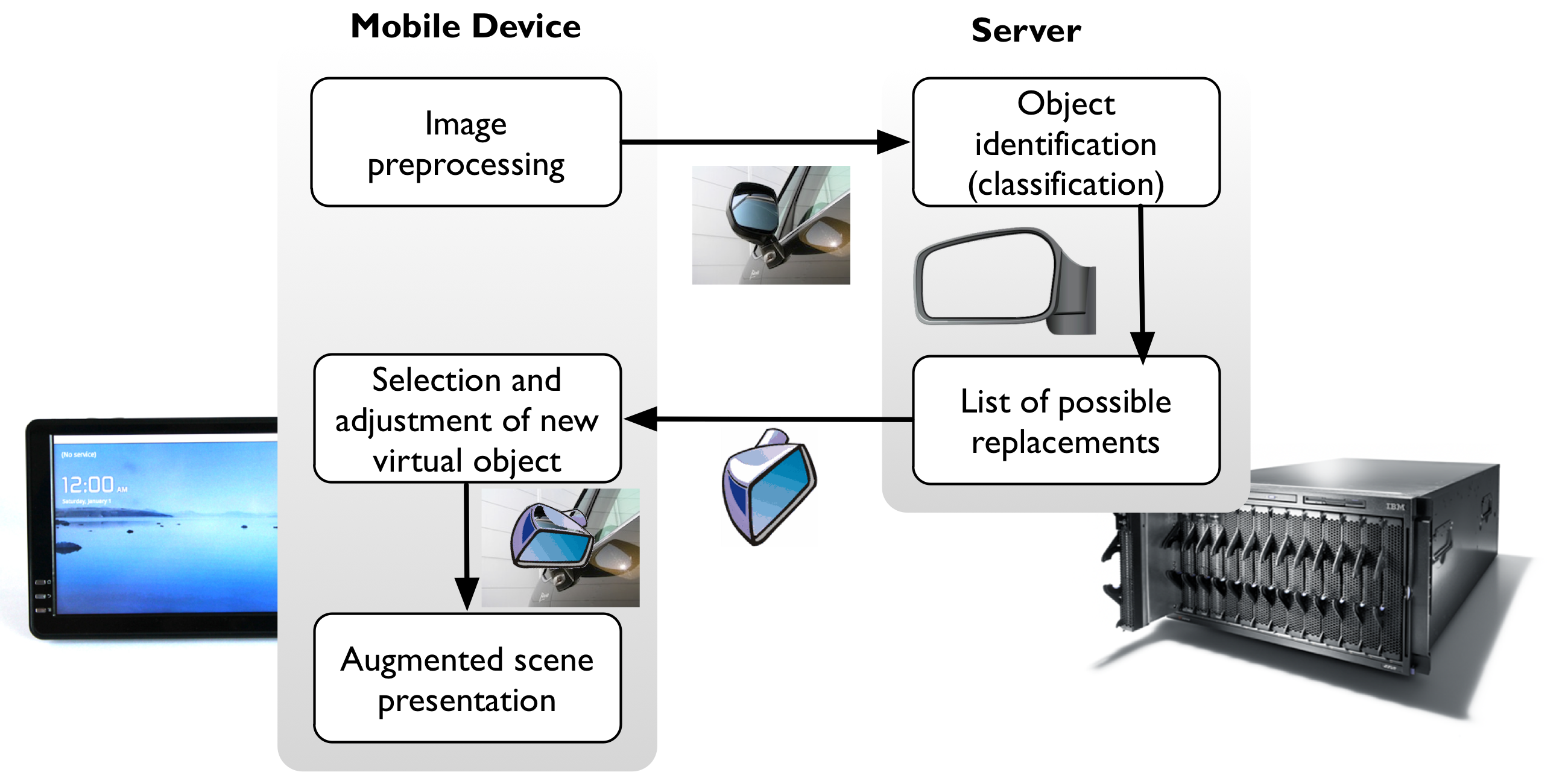}
\end{center}
\caption{General structure of the cloud computing solution for AR}
\label{cloud}
\end{figure}

Therefore we argue, that for advanced mobile AR applications, the cloud computing principle is a significant gain. The client side of the application can compute basic preprocessing and local changes, while the cloud service implements the object recognition technique as outlined on Fig. \ref{cloud}. Furthermore, outlined service must be apropriately designed to receive a re-usable and reliable solution (see \cite{26}).

\section{Conclusion}
In this article we have presented a concept of an augmented reality application targeted to help car designers and customizers. The application is designed for mobile devices capable of creating video composite augmented reality. It uses a neural network recognition algorithm to identify replacable parts on a real vehicle. It also takes advantage of cloud computing principles in that the computationally expensive operations are performed on the server. Also the large database of replacable car parts is stored on the server. Therefore this concept guarantees good scalability and performance.

\vspace*{10pt} \noindent {\bf Acknowledgement:}
This paper is written as a part of a solution of project IGA FBE MENDELU 31/2011 and research plan FBE MENDELU: MSM 6215648904.



\begin{thebibliography}{99}
\vskip12pt

\bibitem{1} Weiser, Mark. Some computer science issues in ubiquitous computing. Communications of the ACM 36, 7, pp. 75-84. ACM, 1993.

\bibitem{2} Billinghurst, Mark and Kato, Hirokazu. Collaborative augmented reality. Communications of the ACM 45, 7, pp. 64-70. ACM, 2002.

\bibitem{3} Kroeker, Kirk L. Mainstreaming augmented reality. Communications of the ACM 53, 7, pp. 19-21. ACM, 2010.

\bibitem{4} Bimber, O. and Raskar, R. Spatial Augmented Reality: Merging Real 
and Virtual Worlds, A.K. Peters, Natick, MA, 2005.

\bibitem{18} Stastny, J. Prochazka, D. Koubek, T. Landa, J.: Augmented reality usage for prototyping speed up. Acta univ. agric. et silvic. Mendel. Brun., 2011, LIX, No. 2. 

\bibitem{16} Ong, S.K. Nee, A.Y.C. and Ong, Soh K.. Virtual Reality and Augmented Reality Applications in Manufacturing. pp. 392, Springer, 2004.

\bibitem{5} Hugues, Olivier Cieutat, Jean-Marc and Guitton, Pascal. An experimental augmented reality platform for assisted maritime navigation. In Proceedings of the 1st Augmented Human International Conference (AH '10). ACM, New York, NY, USA, 2010.

\bibitem{6} Kato, H. Tachibana, K. Billinghurst, M. and Grafe, M. A registration method based on texture tracking using ARToolKit, In Augmented Reality Toolkit Workshop, pp. 77- 85, 2003.



\bibitem{17} Wagner, Daniel Mulloni, Alessandro and Schmalstieg, Dieter. Real-Time Detection and Tracking for Augmented Reality on Mobile Phones, IEEE Transactions on Visualization and Computer Graphics, vol. 16, no. 3, pp. 355-368. IEEE, 2010.

\bibitem{9} Seo, Byung-Kuk Kim, Kangsoo Park, Jungsik and Park, Jong-Il. A tracking framework for augmented reality tours on cultural heritage sites. In Proceedings of the 9th ACM SIGGRAPH Conference on Virtual-Reality Continuum and its Applications in Industry (VRCAI '10), pp. 169-174. ACM, New York, NY, USA, 2010.

\bibitem{19} Vasilev, Julian. The role of DBMS in Analytical Processes of the Logistics of Stocks Processes, International Journal "Information Theories \& Applications", vol. 15, pp. 167-170. FOI Bulgaria, 2008.

\bibitem{10} Ma, Xiaohu Shi, Gang and Tian, Hongbo. Adaptive threshold algorithm for multi-marker augmented reality system. In Proceedings of the 9th ACM SIGGRAPH Conference on Virtual-Reality Continuum and its Applications in Industry (VRCAI '10), pp. 71-74. ACM, New York, 2010. 


\bibitem{12} Bradski, Gary and Kaehler, Adrian. Learning OpenCV: Computer Vision with the OpenCV Library,  pp. 576. O'Reilly Media, 2003.

\bibitem{20}Skorpil, V. and Stastny, J. Comparison Methods for Object Recognition. In Proceedings of the 13th WSEAS International Conference on Systems. Rhodos, GREECE, 2009. pp. 607-610. ISBN 978-960-474-097-0.


\bibitem{24}Skorpil, V. and Stastny, J. Back-Propagation and K-Means Algorithms Comparison. In 2006 8th International Conference on SIGNAL PROCESSING Proceedings. Guilin, CHINA, IEEE Press, 2006. pp. 1871-1874. ISBN 978-0-7803-9736-1.

\bibitem{21}Skorpil, V. and Stastny, J. Comparison of Learning Algorithms. In 24 th Biennial Symposium on Communications. Kingston, CANADA, 2008. pp. 231-234. ISBN 978-1-4244-1945- 6.

\bibitem{28} Wang, Z., Shen, Y., Ong, S. K., and Nee, A. Y.-C. Assembly Design and Evaluation Based on Bare-Hand Interaction in an Augmented Reality Environment. 2009 International Conference on CyberWorlds, pp. 21-28. Ieee. doi: 10.1109/CW.2009.15.


\bibitem{14} Huerta-Canepa, Gonzalo and Lee, Dongman. A virtual cloud computing provider for mobile devices. In Proceedings of the 1st ACM Workshop on Mobile Cloud Computing \& Services: Social Networks and Beyond (MCS '10). ACM, New York, NY, USA, 2010. 

\bibitem{15} Ha, Jaewon Cho, Kyusung and Yang, H. S. Scalable recognition and tracking for mobile augmented reality. In Proceedings of the 9th ACM SIGGRAPH Conference on Virtual-Reality Continuum and its Applications in Industry (VRCAI '10), pp. 155-160. ACM, New York, NY, USA, 2010. 

\bibitem{26} Rychly, Marek and Zendulka, Jaroslav: Modelling of Component-Based Systems with Mobile Architecture. Brno University of Technology, Brno, 2010.




\end{thebibliography}
\end{document}